\documentclass[letterpaper, 10 pt, conference]{ieeeconf}  
\usepackage[table]{xcolor} 
\usepackage[protrusion=false]{microtype}\usepackage{url}
\usepackage{textcomp}

\usepackage{colortbl}

\usepackage{amsmath}
\usepackage{amssymb}
\usepackage{amsfonts}
\usepackage{algorithm}
\usepackage{mathtools}
\usepackage{multirow}
\usepackage{nicefrac}
\usepackage[mathscr]{eucal}
\usepackage{pifont}
\usepackage{bbding}
\usepackage{fontawesome5}
\usepackage{tabularx}
\usepackage{algorithm}
\usepackage{algpseudocode}  
\usepackage{xcolor}
\definecolor{darkred}{rgb}{0.6, 0.0, 0.0}
\usepackage{graphicx}
\usepackage{epsfig}
\usepackage[font=tiny]{subcaption}
\usepackage[export]{adjustbox}
\usepackage{placeins}
\usepackage{float}
\usepackage{forest}
\useforestlibrary{edges}
\usetikzlibrary{shadows.blur}
\graphicspath{{./figures/}}

\usepackage{booktabs}
\usepackage{multirow}
\usepackage{makecell}
\usepackage{arydshln}
\let\labelindent\relax
\usepackage{enumitem}
\usepackage{multicol}
\usepackage{lipsum}
\usepackage{comment}
\usepackage{array}
\usepackage{listings}
\usepackage{pythonhighlight}
\usepackage{tcolorbox}

\tcbset{
    colback=gray!15,
    colframe=gray!40,
    boxrule=0.4pt,
    arc=2pt,
    left=6pt,
    right=6pt,
    top=4pt,
    bottom=4pt,
}

\lstdefinestyle{mypython}{
    language=Python, 
    basicstyle=\ttfamily\small, 
    keywordstyle=\color{blue}, 
    commentstyle=\color{gray}, 
    stringstyle=\color{red}, 
    showstringspaces=false, 
    breaklines=true, 
    frame=none
}

\usepackage{listings}
\usepackage{xcolor}

\definecolor{kw}{RGB}{0, 112, 26}       
\definecolor{fn}{RGB}{0, 60, 160}        
\definecolor{str}{RGB}{163, 21, 21}      
\definecolor{cmt}{RGB}{130, 130, 130}    
\definecolor{bg}{RGB}{249, 249, 249}     
\definecolor{bestgreen}{rgb}{0.0,0.6,0.0}
\definecolor{worstred}{rgb}{0.8,0.0,0.0}

\lstnewenvironment{pythonic}{
  \lstset{
    language=Python,
    basicstyle=\ttfamily\small,
    keywordstyle=\color{kw}\bfseries,
    stringstyle=\color{str},
    commentstyle=\color{cmt}\itshape,
    emphstyle=\color{fn}\bfseries,
    emph={mimic_agent, motion_planner, coder_agent, code_executor,
          unit_test, motion_diagnosis, code_modifier},
    numberstyle=\tiny\color{cmt},
    numbers=left,
    numbersep=8pt,
    backgroundcolor=\color{bg},
    frame=single,
    rulecolor=\color{black!20},
    framesep=6pt,
    xleftmargin=18pt,
    framexleftmargin=18pt,
    showstringspaces=false,
    breaklines=true,
    aboveskip=4pt,
    belowskip=4pt,
    tabsize=2,           
    basewidth=0.5em,     
  }
}{}
\definecolor{darkgreen}{rgb}{0., 0.85, 0.5}
\definecolor{green}{rgb}{0.0, 0.5, 0.0}

  \def\0{{\bf 0}} \def\1{{\bf 1}}

\makeatletter
\newcommand\fs@spacedruled{%
    \def\@fs@cfont{\bfseries}%
    \let\@fs@capt\floatc@ruled
    \def\@fs@pre{\kern1.5mm\hrule height.8pt depth0pt \kern2pt}%
    \def\@fs@mid{\kern2pt\hrule\kern2pt}%
    \def\@fs@post{\kern2pt\hrule\relax}%
    \let\@fs@iftopcapt\iftrue
}
\floatstyle{spacedruled}
\restylefloat{algorithm}
\makeatother
\usepackage{hyperref}

\usepackage[capitalize,noabbrev]{cleveref}

\IEEEoverridecommandlockouts                              
\title{\LARGE \bf
MimicAgent: Quadruped Skills via Prompt-to-Trajectory Generation
}

\author{Lucky Kant Nayak$^*$, Narayanan Palghat Parameswaran$^*$, Neehar Peri, Deva Ramanan \\ Carnegie Mellon University}

\begin{document}

\maketitle
\thispagestyle{empty}
\pagestyle{empty}


\begin{abstract}

We present MimicAgent, a prompt-to-trajectory generation framework for learning dynamic quadruped skills. Although reward shaping is extensively used when training quadruped policies, navigating the resulting reward landscape is notoriously difficult, requiring hours of ``graduate student descent''. Eureka attempts to automate reward design with LLMs, but we find that it struggles to generalize across diverse skills and morphologies. Our key observation is that it is far easier for a human -- and by association, an LLM -- to generate reference motions than to shape reward functions. Our hypothesis is motivated by the success of example-guided RL for humanoids, which exploits large-scale motion capture datasets as references for training locomotion policies. Unlike humanoids, quadrupeds lack such reference motion data. Towards this end, we propose MimicAgent, an agentic harness that, given a skill prompt, generates quadruped reference trajectories with coding agents. These coarse reference trajectories are then used to train example-guided RL policies that are deployable in simulation and in the real-world. Notably, we find that when prompting Claude Fable 5.1 within our agentic harness, 87\% of prompts yield semantically aligned reference trajectories. Please see our \href{https://luckykantnayak.github.io/mimic-agent/}{project page} for accompanying videos.
\end{abstract}

\section{Introduction}
In recent years, learning-based locomotion policies have demonstrated impressive agility and robustness across a wide range of behaviors \cite{margolis2023walk} and terrains \cite{lee2020learning}. However, training such RL policies typically relies on carefully tuned rewards. For complex and highly dynamic skills, designing bespoke reward functions is notoriously difficult and often requires extensive trial-and-error \cite{kumar2021rma}. This challenge has motivated recent efforts to eliminate manual reward engineering entirely.

\textbf{LLMs for Reward Design.}
Recent work has explored using LLMs to automate reward design \cite{yu2023language, ma2023eureka, cui2025grove, turcato2025towards}. For example, Eureka \cite{ma2023eureka} leverages coding LLMs to generate and iteratively refine reward functions from natural language task descriptions. While promising, these methods still depend on hand-engineered success metrics to evaluate candidate rewards and often struggle to generalize across diverse skills and morphologies. In practice, generated reward functions are brittle and often don't accurately follow natural language guidance, particularly for underexplored robot embodiments such as wheeled quadrupeds (Table \ref{tab:baseline}). 


\begin{figure}[t]
    \centering
    \centering
    \includegraphics[trim={1cm 0.75cm 1cm 1.75cm},clip, width=0.97\linewidth]{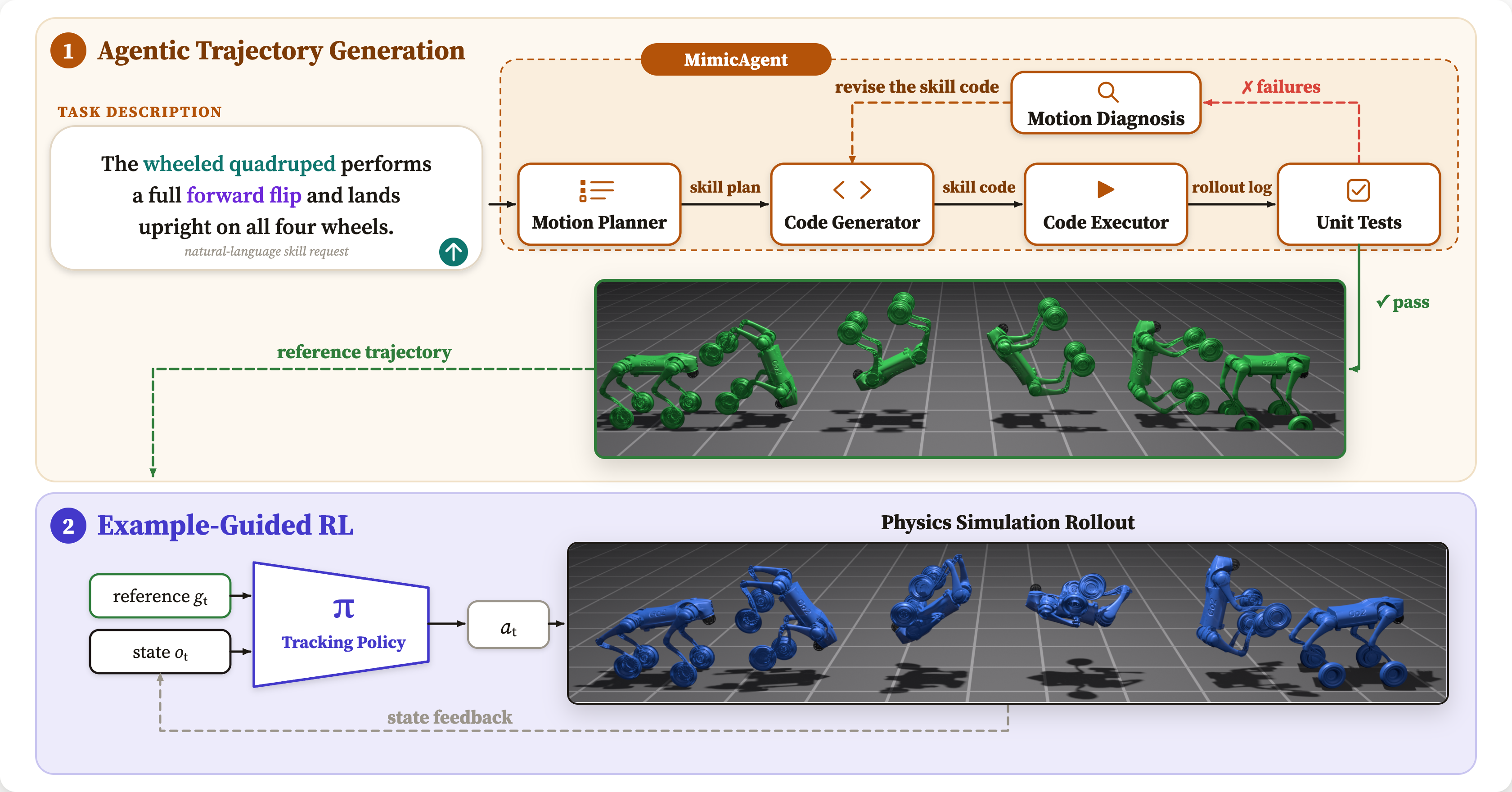}
    \caption{\textbf{Overview.} MimicAgent uses LLM coding agents within a recursive self-improvement loop to generate coarse reference trajectories (shown in \textcolor{green}{green}) from text prompts. Our agentic pipeline converts short text prompts into detailed skill descriptions, translates these into executable functions that describe base and foot motion over time, and outputs them as reference trajectories. We show that these coarse motions are effective reference targets for example-guided RL (with policy rollouts shown in \textcolor{blue}{blue}).}
    \label{fig:teaser}
    \vspace{-8mm}
\end{figure}

\textbf{Learning from Examples.}
An alternative to reward-centric RL is learning from demonstrations. Example-guided RL has been a key enabler for recent progress in humanoid locomotion, where motion capture datasets \cite{AMASS:ICCV:2019, allshire2025visual, xu2025intermimic, he2024learning, peng2018sfv},  or manually annotated keyframes \cite{peng2018deepmimic, zhang2025add} provide rich supervision. However, quadrupeds lack such large-scale, publicly available datasets. Recent works \cite{bellegarda2025allgaits} curate libraries of simple quadruped motion patterns like walking, trotting, and galloping. However, such efforts are orders of magnitude smaller and less diverse than comparable humanoid datasets. This data scarcity limits the applicability of demonstration-based methods despite their strong empirical performance.

\textbf{LLMs for Trajectory Generation.}
To address this gap, we propose MimicAgent, an LLM-based agentic pipeline that directly generates text-guided example trajectories for policy learning. {\em Our key observation is that it is far easier for a human -- and by association, an LLM -- to generate reference motions than to shape reward functions.}
Given a natural language skill description, MimicAgent produces executable code defining base motion and foot trajectories. This code is then executed in a MuJoCo engine using forward kinematics to produce kinematically feasible trajectories, without running a full physics simulator. The resulting trajectories are consumed by a downstream imitation-based RL pipeline like DeepMimic \cite{peng2018deepmimic}.

\textbf{Contributions.} We present three major contributions. 
\begin{enumerate}
    \item \textit{MimicAgent Framework.} We introduce a prompt-to-trajectory agentic harness specifically designed for learning dynamic quadruped skills.
    \item \textit{Learning from Coarse References.} We demonstrate that coarse trajectories generated by MimicAgent are sufficient for training successful sim-to-real policies.
    \item \textit{Trajectory Synthesis vs. Reward Shaping.} Our experiments highlight the limitations of prior LLM-based approaches that rely on reward shaping, and demonstrate the effectiveness of LLM-based trajectory synthesis. 
\end{enumerate}

\section{Related Works}
\textbf{Quadrupedal Locomotion} has matured significantly in recent years \cite{nahrendra2023dreamwaq, smith2023learning}. Early work primarily focused on training legged robots to execute complex contact sequences to achieve different locomotion gaits. A common approach is to use model predictive control (MPC) to track predefined contact patterns \cite{grandia2019feedback}, enabling canonical gaits such as trotting \cite{cheetah3}, pacing \cite{RAIBERT199079}, bounding \cite{eckert_bounding}, and galloping \cite{1255398}. Despite their success, MPC-based approaches require hand-designed reference trajectories and incur high computational costs, limiting behavioral diversity. Recent work extends these ideas to agile bipedal quadrupedal motions \cite{li2024learning} and whole-body loco-manipulation \cite{qiu2025wildlma}. In contrast, our approach leverages example-guided reinforcement learning, which can be more intuitive than designing foot contact sequences.

\textbf{Text-to-Robot Control} is an active area of research for both quadrupeds \cite{zhou2025adaptive, jian2025lapp} and humanoids \cite{cui2024anyskill, cui2025grove}. Early methods relied on structured text templates or NLP tools like parse trees to extract constraints and generate robot trajectories \cite{Kress-Gazit01012008}. Other approaches used representation learning to train language-conditioned policies that map free-form instructions directly to actions \cite{brohan2022rt, mees2022matters}. However, such approaches require large annotated datasets with paired text and robot control, which is difficult to collect for diverse locomotion behaviors. More recent methods prompt LLMs to generate robot code, bridging the gap between language and motor control via intermediate representations like high-level plans, primitive skills, or trajectories \cite{lin2023text2motion, liang2022code, singh2022progprompt}. Inspired by prior work, we leverage LLMs to directly generate reference trajectories for imitation learning.

\section{Dynamic Skills via Trajectory Synthesis}
MimicAgent aims to automate motion imitation learning by replacing manual reward engineering with an agentic motion synthesis pipeline. Rather than navigating a complex and brittle reward landscape, MimicAgent leverages the observation that generating a plausible reference trajectory is often significantly easier than designing a robust reward function for the same behavior. 

\begin{figure*}[!t]
    \vspace{3.mm}
    \centering
    \includegraphics[width=0.95\textwidth]{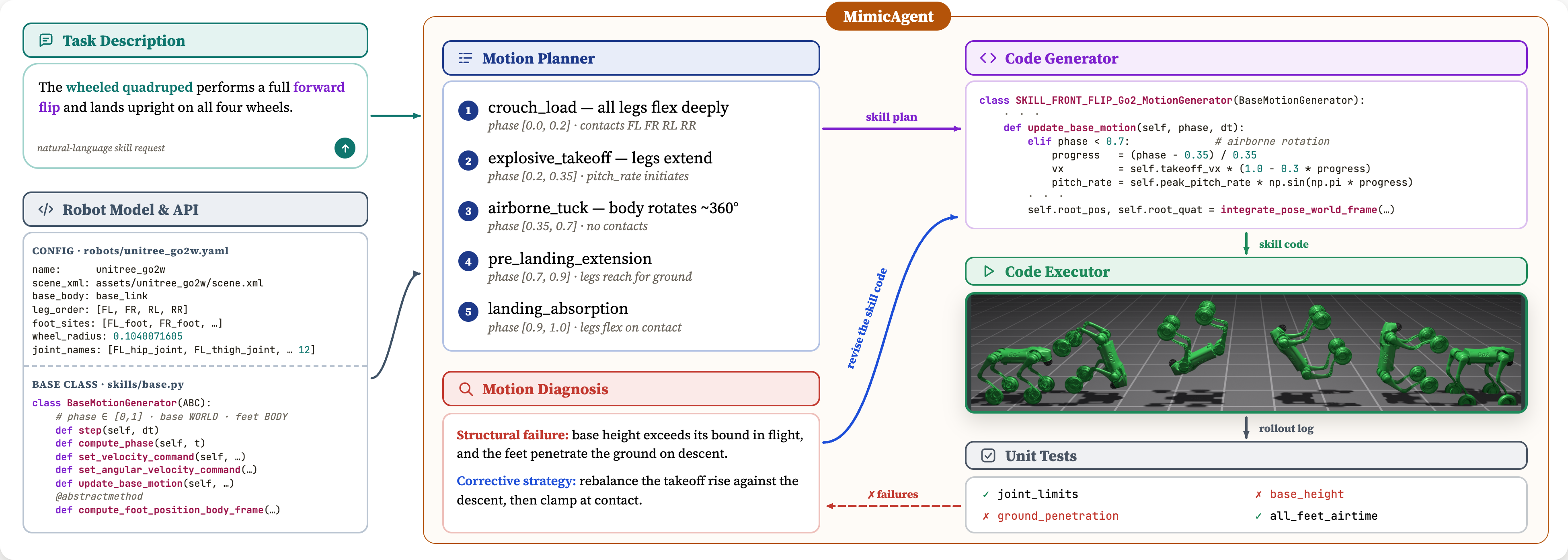}
    \caption{\textbf{MimicAgent Pipeline.} Given the prompt ``the wheeled quadruped performs a full forward flip and lands upright on all four wheels'', our motion planner agent draws on prior knowledge of the robot model and the simulation environment API to decompose the skill into subparts. The code generator agent converts this phase plan into executable base and foot trajectories, which the kinematic executor renders. Task-agnostic unit tests then check the rendered motion for violations. The motion-diagnosis agent summarizes any failed checks and returns them as targeted feedback for code revision. This self-improvement loop repeats until the trajectory passes validation or the iteration budget is exhausted. 
    }
    \label{fig:mimicagent_detailed}
    \vspace{-6mm}
\end{figure*}

\textbf{Agentic Motion Synthesis Pipeline.} 
MimicAgent is an iterative pipeline that decomposes motion synthesis into \textit{motion planning, code generation, execution}, and \textit{refinement} (Figure \ref{fig:mimicagent_detailed}). The input to the system is a natural language skill prompt, 
which is processed by a {\em motion planner} agent to produce a detailed plan encoding the temporal and kinematic structure of the skill. Each motion plan is a textual description of the overall motion intent and stability strategy, the decomposition into sub-phases with associated leg contact states, coordination patterns across leg groups, base linear and angular velocity commands in the body frame, and qualitative leg trajectories relative to the body. It also provides handoff notes for the code generation agent for controller implementation. 


Each skill is parameterized by a normalized phase variable $\phi \in [0,1]$, mapping motions of arbitrary duration into a unified phase-indexed form. A reference trajectory is defined as a sequence of keyframes sampled at discrete phase steps $\phi_t = t/T$, denoted as $\boldsymbol{\tau} = \{s_0, s_1, \ldots, s_T\}$. At each phase step $t$, the keyframe state is: $s_t = [\mathbf{p}_t^{\mathrm{base}},\,
\boldsymbol{\alpha}_t^{\mathrm{base}},\, \boldsymbol{\theta}_t]$, where $\mathbf{p}_t^{\mathrm{base}} \in \mathbb{R}^3$ represents the robot base 3D position, $\boldsymbol{\alpha}_t^{\mathrm{base}} \in \mathbb{R}^4$ is the base orientation represented as a unit quaternion, and $\boldsymbol{\theta}_t \in \mathbb{R}^n$ is the vector of $n$ joint rotations. This representation enables a compact and interpretable parameterization capable of expressing a wide range of locomotion behaviors.

\textbf{Code Generation and Trajectory Execution.}
Given the motion plan, a {\em  code generation} agent synthesizes an executable program that defines the temporal functions and sequence for base and foot trajectories. This generated code is executed via forward kinematics within a MuJoCo simulator. Importantly, we do not perform any physics simulation. As a result, the synthesized trajectories are kinematically consistent but generally violate dynamic constraints. MimicAgent explicitly embraces this tradeoff, delegating dynamic feasibility to the reinforcement learning stage.

\textbf{Task-Agnostic Unit Tests.} To validate generated trajectories without introducing task-specific biases, MimicAgent employs a set of \emph{task-agnostic unit tests}. Rather than optimizing task rewards, MimicAgent filters generated motions using task-agnostic constraints, enabling scalable reference trajectory generation without manually specifying task-specific success criteria. These tests are designed to reject physically implausible or degenerate trajectories while remaining broadly applicable across skills. Each unit test computes a scalar reward, and the aggregate score determines whether a trajectory passes validation. The unit tests include:

\begin{itemize}
    \item \textit{Base Height Envelope.} The base height is constrained to remain within a reasonable envelope relative to the ground. Heights below 0.1~meters or those exceeding twice the standing base height are penalized. This test returns a binary reward of $+1$ or $-1$.
    \item \textit{Joint Violation Penalty.} Joint limit violations are penalized proportional to their frequency. The reward is computed as $r_{\text{joint}} = -1 \times N_{\text{violated}}$, where $N_{\text{violated}}$ denotes the number of joints exceeding their limits.
    \item \textit{Ground Penetration.} Foot-ground penetration is penalized based on severity. Trajectories in which all four feet penetrate the ground incur a large penalty, while partial penetration results in a smaller penalty. The reward is bounded between $+1$ and $-1$.
    \item \textit{Feet Air-Time Constraint.} For non-aerial skills, at least one foot must remain in contact with the ground for half of the motion cycle. This constraint prevents degenerate motions that leave the ground for extended durations.
\end{itemize}

Importantly, these unit tests are skill-agnostic and do not encode task success, performance objectives, or stylistic preferences. They serve as structural validity checks, ensuring that generated trajectories are suitable as reference trajectories.
In contrast, Eureka~\cite{ma2023eureka} relies heavily on carefully designed task-specific success functions to generate reward signals. Defining such success criteria is a time-intensive process that prohibits learning large skill libraries. 

\textbf{Recursive Self Improvement Loop.}
If a trajectory fails validation, a diagnostic agent analyzes the failure in the context of the full system state, including the skill description, motion plan, generated code, and unit test outcomes. The diagnostic agent produces natural language feedback identifying likely failure modes. A code modification agent then revises the trajectory-generating program accordingly. This process iterates until a trajectory passes all unit tests or a fixed iteration budget is reached (N=3). Valid trajectories are returned for downstream imitation learning.

By shifting the focus from reward optimization to reference trajectory synthesis, MimicAgent avoids the brittle and time-consuming process of manual reward engineering. Unlike LLM-based reward design methods that require repeated policy training and hand-crafted evaluation criteria, MimicAgent generates supervision directly in trajectory space. This makes the framework efficient, interpretable, and naturally extensible to diverse quadruped skills.

        


\textbf{Example-Guided RL and Sim-to-Real Transfer.}
Given a coarse reference trajectory, we formulate motion imitation as goal-conditioned reinforcement learning, following DeepMimic~\cite{peng2018deepmimic}. Similar to HOVER~\cite{he2024learning}, we first train a privileged teacher policy with PPO using full simulator state and subsequently distill it into a deployable student policy whose observations can be reconstructed from onboard sensing. At each control step $t$ ($50$~Hz), the policy
$\pi(a_t \mid s_t^p, s_t^g)$ maps the proprioceptive observation $s_t^p$ and reference-goal observation $s_t^g$ to a normalized action vector $a_t \in [-1,1]^{16}$ for the 16-DoF Go2-W ($12$ leg joints and $4$ wheels).  The teacher uses global yaw information available only in simulation, whereas the student employs a yaw-invariant observation representation suitable for hardware deployment. Their self-orientation and reference-orientation inputs are defined as follows:

\begin{align}
s^{p\text{-teacher}}_t &=
\big[\, \mathbf{q}^{\text{root}}_t,\;
\boldsymbol{\omega}^{\text{root}}_t,\;
\boldsymbol{\theta}_t,\;
\dot{\boldsymbol{\theta}}_t,\;
a_{t-1} \,\big], \\
s^{p\text{-student}}_t &=
\big[\, \mathbf{g}_t,\;
\boldsymbol{\omega}^{\text{root}}_t,\;
\boldsymbol{\theta}_t,\;
\dot{\boldsymbol{\theta}}_t,\;
a_{t-1} \,\big],
\end{align}

\vspace{-2mm}
and the goal blocks are
\vspace{-2mm}

\begin{align}
s^{g\text{-teacher}}_t &=
\big[\, \Delta\mathbf{p}^{\text{root}}_t,\;
\hat{\mathbf{q}}^{\text{root}}_t,\;
\hat{\mathbf{v}}_t,\;
\hat{\boldsymbol{\omega}}_t,\;
\hat{\boldsymbol{\theta}}_t,\;
\hat{\dot{\boldsymbol{\theta}}}_t,\;
\Delta\mathbf{f}_t,\;
\phi_t \,\big], \\
s^{g\text{-student}}_t &=
\big[\, \hat{\mathbf{g}}_t,\;
\hat{\mathbf{v}}_t,\;
\hat{\boldsymbol{\omega}}_t,\;
\hat{\boldsymbol{\theta}}_t,\;
\hat{\dot{\boldsymbol{\theta}}}_t,\;
\Delta\mathbf{f}_t,\;
\phi_t \,\big],
\end{align}

where $\mathbf{q}^{\text{root}}_t$ is the root orientation quaternion (sim-only), $\mathbf{g}_t$ is the body-frame gravity direction (projected gravity), $\boldsymbol{\omega}^{\text{root}}_t$ is the base angular velocity, $\boldsymbol{\theta}_t,\dot{\boldsymbol{\theta}}_t$ are the joint positions and velocities, $a_{t-1}$ is the previous action, $\Delta\mathbf{p}^{\text{root}}_t$ and $\Delta\mathbf{f}_t$ are the body-frame root-position and end-effector tracking errors, $\hat{\mathbf{v}}_t,\hat{\boldsymbol{\omega}}_t$ are the reference velocities, and $\phi_t\in[0,1]$ is the cyclic motion phase. Swapping the teacher's absolute-orientation channels for the single yaw-free reference gravity $\hat{\mathbf{g}}_t$ makes the student's observation viable on hardware and free of IMU yaw drift.

The teacher also receives a privileged block $s^{\text{priv}}_t$ of ground-truth base velocity and height, link positions, contact flags, a short future-reference lookahead, and the per-environment domain-randomization parameters. The teacher network is a symmetric actor--critic, where both networks consume $[\,s^{p\text{-teacher}}_t, s^{g\text{-teacher}}_t, s^{\text{priv}}_t\,]$. In contrast, the student actor only sees the deployable channels plus a $50$-step proprioceptive history $H$,  while its critic reuses $s^{\text{priv}}_t$. 

Once the teacher network converges, it supervises the student via DAgger~\cite{pmlr-v15-ross11a}. The student rolls out $\pi_{\text{student}}(a_t\mid s^{p\text{-student}}_t, s^{g\text{-student}}_t)$, and the oracle $\pi_{\text{teacher}}(\hat{a}_t\mid s^{p\text{-teacher}}_t, s^{g\text{-teacher}}_t)$ is queried on the matched privileged state for target actions $\hat{a}_t$. We behavior-clone ($\lVert a_t-\hat{a}_t\rVert^2$) on the deployable observation. Both stages use domain randomization and observation noise for robust sim-to-real transfer.

The final reward sums imitation terms, each an exponential tracking kernel $\exp(-k\,e)$, and hardware-aware regularization terms:

\begin{equation}
r_t =
\underbrace{\sum_{m\in\mathcal{I}} w_m \exp\!\big(-k_m\, e^m_t\big)}_{\text{imitation}}
\;+\;
\underbrace{\sum_{n\in\mathcal{R}} w_n\, c^n_t}_{\text{regularization}},
\end{equation}

where $e^m_t$ is the squared deviation between simulated and reference quantity for $m\in\{$joint position, joint velocity, root position, root rotation, root linear velocity, root angular velocity, end-effector$\}$. 
The regularization terms $c^n_t$ penalize action rate, joint velocity, joint acceleration, hinge violations of the
joint-position, velocity, and torque limits.

\section{Experiments}
We first describe our evaluation metrics and benchmark our baselines. We then ablate the quality of generated reference motions, comparing human-annotated keyframes against trajectories produced by our agentic synthesis pipeline, and measure the annotation time each requires. Further, we compare the policies learned by our model against hand-tuned RL expert policies for trotting and bounding. Lastly, we ablate the impact of unit tests and iterative refinement on our generated references.

\textbf{Metrics.} We quantitatively compare manual keyframing, Eureka \cite{ma2023eureka}, and MimicAgent with a 57 person user study conducted through Cint. Participants were presented with pairs of textual skill prompts and videos of trained policies corresponding to that task and were asked to rate how well the video of policy rollouts follows the prompt on a 5-point Likert scale. Videos were presented in a randomized order to eliminate ordering bias and participants were not informed which method was used to generate each video. In addition, we compute the mean squared error between the reference trajectories and the trajectories produced by the learned policy for both human annotated keyframes and MimicAgent's keyframes to measure how faithfully the learned policy follows the demonstration. 

\textbf{Baselines.} We compare MimicAgent against Eureka~\cite{ma2023eureka} and
DeepMimic~\cite{peng2018deepmimic} policies learned from human-annotated keyframes across
seven skills: Go2 Trot, Go2 Bounding, Go2-W Front Flip (FF), Go2-W Side Flip (SF), Go2-W Aerial Crossover (AC), Go2-W Reverberating Yaw Pulse (RYP), and Go2-W Crab Diagonal Scuttle (CDS), described below:

\begin{itemize}
    \item \textit{UniTree Go2 Trot.} A symmetric diagonal gait alternating front-left and
    rear-right with front-right and rear-left, maintaining continuous ground contact and a
    stable torso while tracking commanded forward velocity.
    \item \textit{UniTree Go2 Bounding.} A dynamic gait where the front legs push off
    together followed by the rear, producing a brief aerial phase and pronounced vertical
    oscillation under forward propulsion.
    \item \textit{UniTree Go2-W Front Flip.} Strong forward pitch momentum carries the robot
    fully airborne through a 360° rotation about its lateral axis, landing upright with
    controlled impact.
    \item \textit{UniTree Go2-W Side Flip.} A rapid roll about the longitudinal axis carries
    the robot fully airborne through a 360° side rotation, recovering to a stable stance.
    \item \textit{UniTree Go2-W Aerial Crossover.} While carving forward on its wheels, the
    robot lifts alternating diagonal leg pairs off the ground, executing rapid mid-air
    crossover motions.
    \item \textit{UniTree Go2-W Reverberating Yaw Pulse.} An oscillatory motion with a
    decreasing amplitude envelope that mimics physical damping. Phase controls the amplitude
    decay and pulse frequency, giving an accelerating--decelerating rhythm visually distinct
    from constant-rate spinning.
    \item \textit{UniTree Go2-W Crab Diagonal Scuttle.} The robot scuttles diagonally with its
    body axis held perpendicular to the travel vector, decoupling orientation from travel
    direction. Phase coordinates cross-body leg sweeps in which front and rear act in
    opposition.
\end{itemize}

\textit{Eureka.} Eureka formulates reward generation as evolutionary search over reward programs. Given a natural-language task description $\ell$, environment source code $\mathcal{M}$, a policy optimization algorithm $\mathcal{A}$, and a task-level fitness function $F$, the LLM samples a population of $K$ candidate reward programs $\{R_1,\ldots,R_K\}$. Each reward program produces a scalar reward and a dictionary of interpretable reward components. For each candidate, Eureka trains a policy $\pi_k=\mathcal{A}(\mathcal{M},R_k)$ and evaluates its performance using $s_k=F(\pi_k)$. At each iteration, the best-performing candidate $R^{*} = \arg\max_k s_k$ is retained and supplied to the LLM together with reward-reflection feedback. This feedback summarizes the task-level fitness and the values of individual reward components at intermediate policy-training checkpoints, enabling the LLM to modify their scales or composition. This constitutes one evolutionary iteration; we repeat for $N$ iterations, optionally with random restarts, and return the highest-fitness program. We use the default configuration with $N=5$ evolutionary-search iterations and $K=16$ candidate rewards per iteration, yielding 80 candidate reward programs per skill.

Eureka's search depends on how accurately $F$ captures the target behavior. Since no predefined evaluation metrics exist for our novel Go2-W skills, we manually construct a task-specific fitness function for each skill. For example, the fitness function for a side flip measures whether the robot completes the desired airborne rotation and recovers to an upright configuration. This provides Eureka with explicit task-success supervision during reward search. In contrast, MimicAgent applies a shared set of structural validation checks across skills and does not optimize a task-specific fitness metric. Despite this favorable evaluation setting, we find that Eureka produces less reliable policies for the stylistically diverse Go2-W behaviors considered in our experiments.

\textit{Keyframing Reference Motion Tool.} To streamline reference-motion animation, we built a custom keyframe annotation tool with Claude that visualizes the  kinematics of a robot model in MuJoCo and lets users ``puppeteer'' the robot to generate reference trajectories. Our tool uses click-and-drag mechanics for coarse spatial adjustment, and a slider-based UI panel for fine-grained control of individual foot positions ($X, Y, Z$) and root body orientation (roll, pitch, yaw). Leg mirroring, a keyframe timeline, and trajectory format selection further accelerate annotation. We argue that this interactive design lowers the barrier to authoring reference motion -- for human operators and LLMs alike -- relative to traditional reward shaping workflows.

\begin{table}[t]
\vspace{1mm}
\centering
\caption{\textbf{Comparison to State-of-the-Art.} We conduct a user study (n=57) to compare the semantic alignment of policies generated with human-annotated keyframes, Eureka, and MimicAgent on a scale of 1 to 5. 
Higher is better.}
\label{tab:baseline}
\resizebox{\linewidth}{!}{
\begin{tabular}{l|cc|ccccc}
\toprule
\textbf{Robot} 
& \multicolumn{2}{c|}{\cellcolor{blue!8}\textbf{Go2}} 
& \multicolumn{5}{c}{\cellcolor{green!8}\textbf{Go2-W}} \\
\cmidrule(lr){2-3} \cmidrule(lr){4-8}
\textbf{Method $\downarrow$\;\;/ Skill $\rightarrow$} 
& \textbf{Trot} & \textbf{Bound} 
& \textbf{SF} & \textbf{FF} & \textbf{AC} & \textbf{CDS} & \textbf{RYP} \\
\midrule
Keyframe Tool
& 2.9 $\pm$ 1.4 & \textbf{3.5 $\pm$ 1.4} & 3.5 $\pm$ 1.2 & 4.2 $\pm$ 1.0 & 2.3 $\pm$ 1.3 & 2.5 $\pm$ 1.2 & 1.8 $\pm$ 1.2 \\
Eureka
& 2.6 $\pm$ 1.5 & 2.8 $\pm$ 1.4 & 1.4 $\pm$ 1.1 & 1.1 $\pm$ 0.3 & 1.9 $\pm$ 1.1 & 2.1 $\pm$ 1.1 & \textbf{4.3 $\pm$ 1.4} \\
{MimicAgent}
& \textbf{4.4 $\pm$ 1.0} & 3.2 $\pm$ 1.4 & \textbf{4.9 $\pm$ 0.3} & \textbf{4.6 $\pm$ 0.8} & \textbf{4.2 $\pm$ 1.0} & \textbf{4.5 $\pm$ 0.5} & 2.6 $\pm$ 1.4 \\
\bottomrule
\end{tabular}
}
\end{table}

\begin{table}[t]
\centering
\caption{\textbf{Comparison of Time-to-Annotation.} We evaluate the time to annotation (in minutes) between human-annotated keyframes and MimicAgent keyframes. 
Lower is better.}
\label{tab:time}
\resizebox{\linewidth}{!}{
\begin{tabular}{l|cc|ccccc}
\toprule
\textbf{Robot} 
& \multicolumn{2}{c|}{\cellcolor{blue!8}\textbf{Go2}} 
& \multicolumn{5}{c}{\cellcolor{green!8}\textbf{Go2-W}} \\
\cmidrule(lr){2-3} \cmidrule(lr){4-8}
\textbf{Method $\downarrow$\;\;/ Skill $\rightarrow$} 
& \textbf{Trot} & \textbf{Bound} 
& \textbf{SF} & \textbf{FF} & \textbf{AC} & \textbf{CDS} & \textbf{RYP} \\
\midrule
Keyframe Tool
& 2.6 & 4.5 
& 4.5 & 4.4 & 7.6 & 14.5 & 12.4 \\
{MimicAgent}
& \textbf{1.1} & \textbf{1.2} 
& \textbf{2.9} & \textbf{3.2} & \textbf{3.6} & \textbf{1.6} & \textbf{1.8} \\
\bottomrule
\end{tabular}
}
\vspace{-6mm}
\end{table}

\begin{table}[t]
\vspace{1mm}

\centering
\caption{\textbf{Comparison of Reference Motion Quality.} We evaluate the tracking error of policies trained with human-annotated  and MimicAgent keyframes. Across all skills and metrics, MimicAgent policies achieve lower mean squared error on both pose-level errors (root and body position and rotation) and motion-level errors (linear and angular velocities).
Lower is better.
}
\label{tab:reference}
\resizebox{\linewidth}{!}{
\begin{tabular}{l c|cc|ccccc}
\toprule
& \textbf{Robot} 
& \multicolumn{2}{c|}{\cellcolor{blue!8}\textbf{Go2}} 
& \multicolumn{5}{c}{\cellcolor{green!8}\textbf{Go2-W}} \\
\cmidrule(lr){3-4} \cmidrule(lr){5-9}
\textbf{Method} & \textbf{Metric $\downarrow$\;\;/ Skill $\rightarrow$} 
& \textbf{Trot} & \textbf{Bound} 
& \textbf{SF} & \textbf{FF} & \textbf{AC} & \textbf{CDS} & \textbf{RYP} \\
\midrule

\multirow{6}{*}{Keyframe Tool}
& $E_{\text{RP}}$
& 0.170 & 0.076 
& 0.199 & 0.099 & {0.020} & 0.044 & 0.041 \\
& $E_{\text{RR}}$
& 0.177 & 0.208 
& 0.238 & 0.208 & {0.025} & 0.201 & 0.142 \\
& $E_{\text{JR}}$
& 0.360 & 0.320 
& 0.484 & \textbf{0.507} & 0.396 & 0.033 & 0.025 \\
& $E_{\text{JV}}$
& 4.313 & 3.868 
& 8.380 & 15.183 & 6.823 & 6.171 & 4.721 \\
& $E_{\text{LV}}$
& 0.251 & 0.214 
& 0.548 & 0.287 & \textbf{0.035} & 0.261 & 0.126 \\
& $E_{\text{AV}}$
& 0.918 & 0.517 
& 1.883 & \textbf{1.126} & 0.233 & {0.797} & 1.172 \\
\midrule

\multirow{6}{*}{{MimicAgent}}
& $E_{\text{RP}}$
& \textbf{0.052} & \textbf{0.049} 
& \textbf{0.077} & \textbf{0.089} & \textbf{0.013} & \textbf{0.035} & \textbf{0.013} \\
& $E_{\text{RR}}$
& \textbf{0.046} & \textbf{0.114} 
& \textbf{0.207} & \textbf{0.160} & \textbf{0.024} & \textbf{0.032} & \textbf{0.089} \\
& $E_{\text{JR}}$
& \textbf{0.109} & \textbf{0.224} 
& \textbf{0.382} & 0.528 & \textbf{0.373} & \textbf{0.009} & \textbf{0.018} \\
& $E_{\text{JV}}$
& \textbf{0.646} & \textbf{1.580} 
& \textbf{5.299} & \textbf{12.871} & \textbf{4.761} & \textbf{1.609} & \textbf{2.348} \\
& $E_{\text{LV}}$
& \textbf{0.093} & \textbf{0.136} 
& \textbf{0.295} & \textbf{0.264} & 0.044 & \textbf{0.124} & \textbf{0.047} \\
& $E_{\text{AV}}$
& \textbf{0.176} & \textbf{0.353} 
& \textbf{1.869} & 1.697 & \textbf{0.211} & \textbf{0.264} & \textbf{0.455} \\
\bottomrule
\end{tabular}
}
\vspace{-6mm}
\end{table}

\textbf{Comparison with State-of-the-Art.} We conduct a user study (n=57) comparing trained policies generated from human annotated keyframes, Eureka, and MimicAgent. Participants consistently prefer MimicAgent policies for Trotting (+1.8 vs. Eureka), Side Flip (+3.5 vs. Eureka), Front Flip (+3.5 vs. Eureka), Aerial Crossover (+2.3 vs. Eureka), and Crab Diagonal Scuttle  (+2.4 vs. Eureka). For Bounding, users favor policies generated from human-annotated keyframes, though all approaches received relatively low ratings. Eureka achieves the highest score on Reverberating Yaw Pulse (+1.7 vs. MimicAgent). Notably, Eureka still relies on carefully specified task-specific success metrics and exhibits limited robustness when scaling to diverse, stylistically different skills. Therefore, we hand-engineer success criteria for Eureka's policies to give it a better chance of succeeding. Despite this, performance degrades for many wheeled quadruped skills, likely due to the increased joint space complexity. 

\textbf{Time-to-Annotation.}
We compare the time required for generating human-annotated keyframes and MimicAgent generated keyframes in Table \ref{tab:time}. We find that MimicAgent consistently requires less time than manually annotated keyframes. The first five skills (trot, bounding, side flip, front flip, aerial crossover) are intentionally limited to relatively simple motions (e.g., unidirectional velocity or in-place yaw). In contrast, crab diagonal scuttle and reverberating yaw pulse represent more complex skills that incorporate asymmetric leg movements coupled with base rotations, which are time-consuming to keyframe manually. We posit that as skill complexity increases, human annotation time will grow significantly faster than MimicAgent’s skill generation time.

\textbf{Comparison of Reference Motion Quality.} 
We compare DeepMimic policies trained on reference trajectories with 150 frames generated by our manual keyframing tool and MimicAgent in Table \ref{tab:reference}. Policies trained with MimicAgent reference trajectories consistently achieved lower tracking error across almost all skills and evaluation metrics, suggesting that the generated reference trajectories were smoother than the human-annotated trajectories. We track errors for root position in the global frame ($E_{RP}$, $m$), root rotation ($E_{RR}$, rad), joint rotation ($E_{JR}$, $rad$), joint velocity ($E_{JV}$, $rad/s$), 
root linear velocity ($E_{LV}$, $m/s$), and root angular velocity ($E_{AV}$, $rad/s$). This improvement is most pronounced for highly dynamic skills such as side flips, front flips, and aerial crossovers, where policies trained on manually keyframed trajectories exhibit significantly larger velocity and angular velocity errors. In contrast, policies trained on MimicAgent references result in improved motion timing and smoother rotational dynamics during policy learning. Even for simpler skills such as trotting and bounding, MimicAgent yields lower global pose error, reflecting higher reference motion quality. These results indicate that improved reference trajectory quality directly reduces downstream imitation error, demonstrating the effectiveness of MimicAgent as a scalable alternative to manual keyframing for imitation-based RL.

\textbf{Comparison with Expert RL Controller.}
To test whether MimicAgent policies reach the locomotion quality of hand-tuned expert controllers, we compare against Walk-These-Ways (WTW)~\cite{margolis2023walk} on two canonical gaits: trot and bound. However, since these two methods are not guaranteed to produce gaits at the same frequency, their trajectories are not directly comparable. We therefore adopt Waveform Morphology Analysis (WMA), normalizing each trajectory to a single gait cycle for frequency-invariant comparison (Table~\ref{tab:wtw_comparison}). On trot, we find that MimicAgent tracks a $1.0$~m/s command at $0.9999$~m/s, a $6\times$ lower MAE than WTW, while attaining higher periodic stability ($0.958$ vs.\ $0.865$). Its stance duty cycle of $49.8\%$ closely matches the canonical $50/50$ trot definition. Cycle duration varies by only $\pm 0.2$ frames against $\pm 0.5$ for WTW, confirming a more metronomic gait. On bound, MimicAgent tracks the commanded velocity with $30\%$ lower error and again achieves higher periodic stability ($0.959$ vs.\ $0.863$). Notably, MimicAgent spontaneously discovers a more dynamic gait topology; it's $39.4\%$ stance duty cycle (vs.\ $49.8\%$) suggests a genuine aerial flight phase that WTW does not exhibit.
 
\begin{table}[t]
\vspace{1mm}

\centering
\footnotesize
\setlength{\tabcolsep}{3pt}
\caption{
    \textbf{Comparison with Walk-These-Ways.} We compare the gait quality of an RL policy learned from MimicAgent's reference with an expert-tuned controller at a commanded velocity of 1 m/s. MimicAgent achieves $6\times$ lower velocity error on trot and $30\%$ lower on bound.
}
\label{tab:wtw_comparison}
\renewcommand{\arraystretch}{1.1}
\begin{tabular*}{\linewidth}{@{\extracolsep{\fill}} ll|c|c @{}}
\toprule
\textbf{Gait} & \textbf{Metric}
  & \cellcolor{blue!8}\textbf{MimicAgent} (Ours)
  & \cellcolor{green!8}\textbf{WTW} (Expert)\\

\midrule
\multirow{6}{*}{Trot}
  & Mean Vel. (m/s)        & $\mathbf{0.9999}$ & $1.1959$ \\
  & Vel. Error (MAE)       & $\mathbf{0.0351}$ & $0.1959$ \\
  & Periodic Stab.         & $\mathbf{0.9575}$ & $0.8646$ \\
  & Stance Duty (\%)       & $49.8$ & $47.0$ \\
  & Cycle Dur. (Frames)    & $10.0 \pm 0.2$ & $16.6 \pm 0.5$ \\
\midrule
\multirow{6}{*}{Bound}
  & Mean Vel. (m/s)        & $\mathbf{1.0053}$ & $1.1930$ \\
  & Vel. Error (MAE)       & $\mathbf{0.1495}$ & $0.2139$ \\
  & Periodic Stab.         & $\mathbf{0.9587}$ & $0.8625$ \\
  & Stance Duty (\%)       & $39.4$ & $47.7$ \\
  & Cycle Dur. (Frames)    & $9.8 \pm 0.8$ & $16.7 \pm 0.5$ \\
\bottomrule
\end{tabular*}
\end{table}

\textbf{Compute and Token Analysis.}
Tab.~\ref{tab:compute_tokens} compares the compute and token usage of Eureka and MimicAgent when training the privileged-teacher network. Eureka evaluates 80 reward candidates per skill across 5 evolutionary search iterations, whereas MimicAgent trains a reference-tracking teacher under domain randomization for sim-to-real robustness. We exclude student distillation from this comparison, as it is a separate deployment-specific stage and Eureka's policies are not deployable as trained. We find that the two methods use comparable training compute ($46.0$ vs.\ $50.9$ GPU-hours), while Eureka consumes roughly $42\times$ more LLM tokens for reward generation and refinement. However, videos on our project page show that Eureka learns stable locomotion skills yet rarely acquires aerial or parkour skills. Increasing its PPO budget would optimize the selected reward more thoroughly, but cannot correct an objective that fails to encode the intended behavior. We therefore argue that reward-candidate discovery, rather than training or token budget, is Eureka's primary bottleneck on our evaluated skills.

\begin{table}[t]
    \centering
    \caption{\textbf{Compute and Token Comparison.} We compare the compute and token usage between MimicAgent and Eureka below using $8\times$ NVIDIA RTX~3090 GPUs. We average results over 7 skills, and compute token usage with Claude Sonnet 4.5.
    }
    \label{tab:compute_tokens}
    \footnotesize
    \setlength{\tabcolsep}{3pt}
    \renewcommand{\arraystretch}{1.12}
    \begin{tabularx}{\columnwidth}{
        @{}l
        >{\centering\arraybackslash}X
        >{\centering\arraybackslash}X@{}}
        \toprule
        \textbf{Metric}
        & \textbf{Eureka}
        & \textbf{MimicAgent} \\
        \midrule

        Parallel Envs.
        & 4,096
        & 32,768 \\

        PPO Iterations
        & 3,000
        & 10,000 \\

        Avg. Wall Clock Time
        & 5\,h 45\,min
        & 6\,h 22\,min \\

        Avg. GPU Hours
        & 46.0
        & 50.9  \\

        Avg. LLM Tokens
        & 727K
        & 17K \\
        \bottomrule
    \end{tabularx}
    \vspace{-6mm}
\end{table}

\begin{figure}[b]
    \centering
    \includegraphics[width=0.97\linewidth, trim={3cm 0cm 32cm 5cm}, clip]{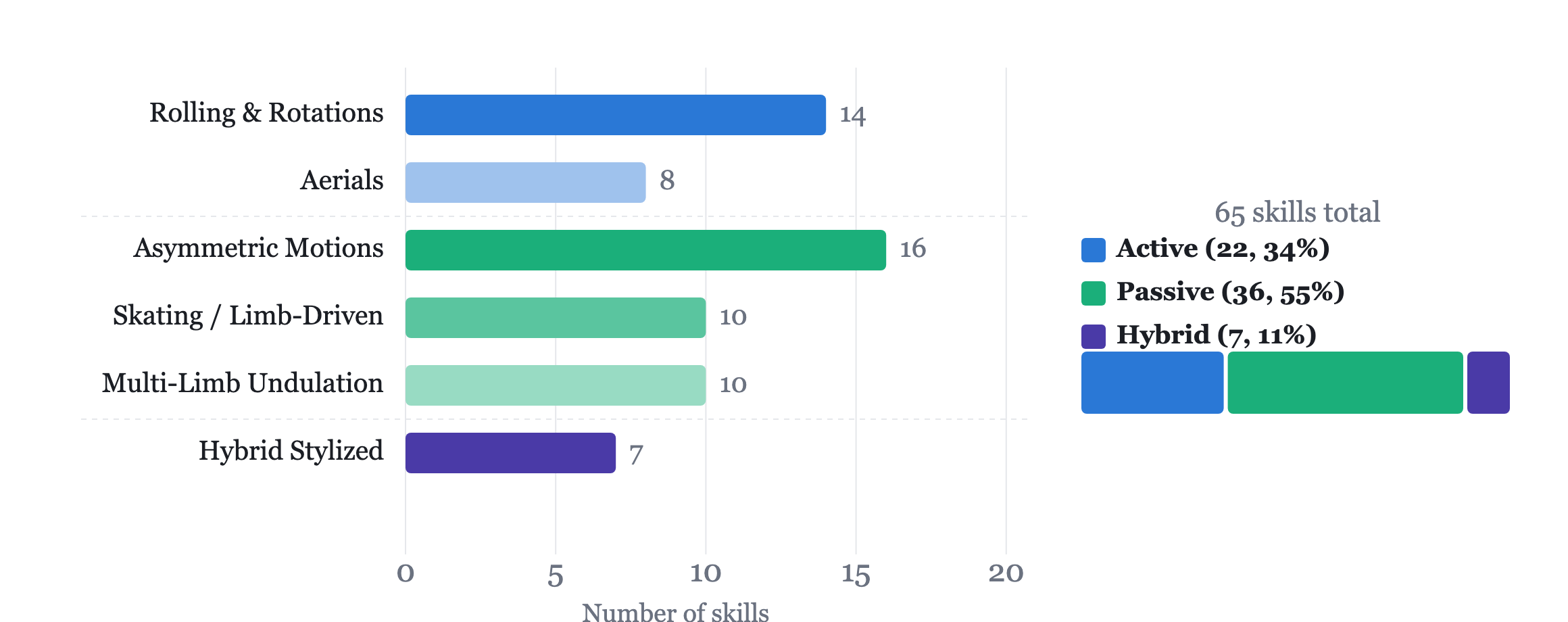}
    \caption{\textbf{LLM Generated Skill Diversity.} MimicAgent generates 65 skills distributed across all six taxonomy categories, covering both 22 Active (wheel-driven), 36 Passive (limb-driven), and 7 Hybrid motion regimes. The balanced spread demonstrates that the pipeline does not collapse onto a narrow subset of the motion space.}
    \label{fig:taxonomy_diversity}
\end{figure}

\textbf{Automatic Skill Generation.} In addition to manually prompting MimicAgent, we can leverage LLMs to automatically generate large-scale skill libraries. We employ a dedicated skill proposal agent that is instructed to propose behaviors that differ in contact patterns, base motion (translation, rotation, or vertical motion), limb coordination, and stylistic or agility characteristics. To prevent duplication, we append all previously generated skills to the prompt context on subsequent iterations, repeating until we reach a target number of unique skills. MimicAgent then processes each description independently. As shown in Fig.~\ref{fig:taxonomy_diversity}, we generated 65 skills spanning six taxonomy categories that cover the Active (wheel-driven), Passive (limb-driven), and Hybrid motion regimes of a wheeled quadruped.

\begin{table}[t]
\vspace{1mm}

\centering
\caption{\textbf{Direct Prompting vs MimicAgent Harness.} We compare unit test success rates between direct code generation and the MimicAgent critic-repair pipeline on 65 LLM proposed skills. Loop~$i$ reports cumulative success after $i$ rounds of critic-driven self-repair. SA reports the human verified semantic alignment of generated reference trajectory with the skill prompt.}
\label{tab:mimicagent-comparison}
\scriptsize
\setlength{\tabcolsep}{5pt}
\begin{tabular}{llccccc}
\toprule
\textbf{Model} & \textbf{Method} & \textbf{No Loop} & \textbf{Loop 1} & \textbf{Loop 2} & \textbf{Loop 3} & \textbf{SA} \\
\midrule
\multirow{2}{*}{Sonnet 5} & Direct     & 54\% & --  & --  & --  & 49\% \\
                          & MimicAgent & 54\% & 74\% & 78\% & 79\% & \textbf{65\%} \\
\midrule
\multirow{2}{*}{Opus 5}   & Direct     & 49\% & --  & --  & --  & 36\% \\
                          & MimicAgent & 68\% & 80\% & 91\% & 93\% & \textbf{82\%} \\
\midrule
\multirow{2}{*}{Fable 5.1} & Direct     & 71\% & --  & --  & --  & 68\% \\
                                  & MimicAgent & 73\% & 91\% & 94\% & 95\% & \textbf{87\%} \\
\bottomrule
\end{tabular}
\vspace{-4mm}
\end{table}

\textbf{Success Rate and Semantic Alignment.} Table.~\ref{tab:mimicagent-comparison} compares three LLMs under Direct (zero-shot prompt-to-code) and MimicAgent-based trajectory generation. \textit{No Loop} reports the unit-test success rate of the first attempt. We find that MimicAgent improves over Direct prompting even without iterative improvement, suggesting that the motion planner agent improves generation quality. Further, \textit{Loop~$i$} reports the cumulative unit-test success rate after $i$ rounds of critic-driven repair. However, passing the automated unit tests does not guarantee semantic correctness. We therefore report semantic alignment (SA) to measure the fraction of all skills whose generated trajectory matches its prompt based on human verification. Even with stronger frontier models, the MimicAgent harness considerably improves unit test success rates and semantic alignment over the baseline.

\textbf{Impact of Unit Tests on Success Rate.}
We ablate the impact of task-agnostic unit tests on the overall success rate of trajectories generated by MimicAgent in Table \ref{tab:human_scores_avg}. To isolate the contribution of different constraints, we define six progressive unit test configurations: no unit tests (C0), adding a base height limit (C1), adding joint limit checks (C2), adding a feet airtime penalty (C3), adding ground penetration checks (C4, default used in the main paper), and adding dynamically generated, skill-specific unit tests proposed by an LLM (C5). To evaluate these configurations, we randomly sample five skills from Sonnet 5's 51 successful skills (79\% of 65 skills) passed by MimicAgent Sonnet. For each skill and unit test configuration, we manually validate (e.g. pass / fail) whether the executed skill closely matches the text description. The final success metric for each configuration is the total success rate across the five evaluated skills.

Our results show that progressively adding unit tests improves the overall success rate, with C4 and C5 achieving the highest policy success rate of 80\%. The dip observed at C3 can be explained by an interaction between constraints. Specifically, enforcing feet airtime in isolation successfully detects when all feet leave the ground, but compensates by driving the body downward to restore contact. Without a ground penetration constraint, this leads to feet clipping into the terrain. This issue is fully resolved at C4, where the ground penetration constraint is introduced alongside the airtime constraint, demonstrating that these two constraints are complementary and must be applied together to be effective. At C5, LLM-proposed tests introduce skill-specific constraints tailored to the nuances of each motion, resulting in a modest but consistent improvement.

\begin{table}[t]
\vspace{1mm}

\centering
\caption{\textbf{Unit Test Ablation.} We ablate the impact of each unit test on reference-trajectory and learned-policy quality below, where checkmarks denote the active tests in each configuration. We find that enabling all unit tests performs best, reaching a 92\% overall prompt-to-policy alignment.}
\label{tab:human_scores_avg}

\renewcommand{\arraystretch}{1.15}
\setlength{\tabcolsep}{2pt}

\newcommand{\yesicon}{%
\textcolor{bestgreen}{%
\raisebox{0.08ex}{\scalebox{1.15}{\(\boldsymbol{\checkmark}\)}}}}

\newcommand{\noicon}{%
\textcolor{worstred}{%
\raisebox{0.03ex}{\scalebox{1.15}{\(\boldsymbol{\times}\)}}}}

\resizebox{\columnwidth}{!}{%
\begin{tabular}{l|ccccc|cc}
\toprule
\textbf{Config}
& \multicolumn{5}{c|}{\cellcolor{green!8}\textbf{Active Unit Tests}}
& \multicolumn{2}{c}{\cellcolor{blue!8}\textbf{Human Score}}
\\

\cmidrule(lr){2-6}
\cmidrule(lr){7-8}

& \textbf{Height}
& \textbf{Joint}
& \textbf{Airtime}
& \textbf{Ground}
& \textbf{LLM}
& \textbf{Reference}
& \textbf{Policy}
\\

\midrule

C0 & \noicon & \noicon & \noicon & \noicon & \noicon & 1/5 & 1/5 \\
C1 & \yesicon & \noicon & \noicon & \noicon & \noicon & 3/5 & 2/5 \\
C2 & \yesicon & \yesicon & \noicon & \noicon & \noicon & 4/5 & 3/5 \\
C3 & \yesicon & \yesicon & \yesicon & \noicon & \noicon & 4/5 & 2/5 \\
C4 & \yesicon & \yesicon & \yesicon & \yesicon & \noicon
   & \textbf{5/5} & \textbf{4/5} \\

\rowcolor{gray!10}
C5 & \yesicon & \yesicon & \yesicon & \yesicon & \yesicon
   & \textbf{5/5} & \textbf{4/5} \\

\bottomrule
\end{tabular}%
}

\vspace{1mm}

\resizebox{\columnwidth}{!}{%
\begin{tabular}{c}
\textbf{RL Policy Semantic Alignment Success Rate: 47/51 (92\%)} \\
\end{tabular}%
}

\vspace{-6mm}
\end{table}

\begin{figure}[t]
     \vspace{2mm}

    \centering
    \begin{subfigure}[t]{0.98\linewidth}
        \centering
        \includegraphics[width=0.98\linewidth, trim={0 1cm 0 2cm},  
        clip]{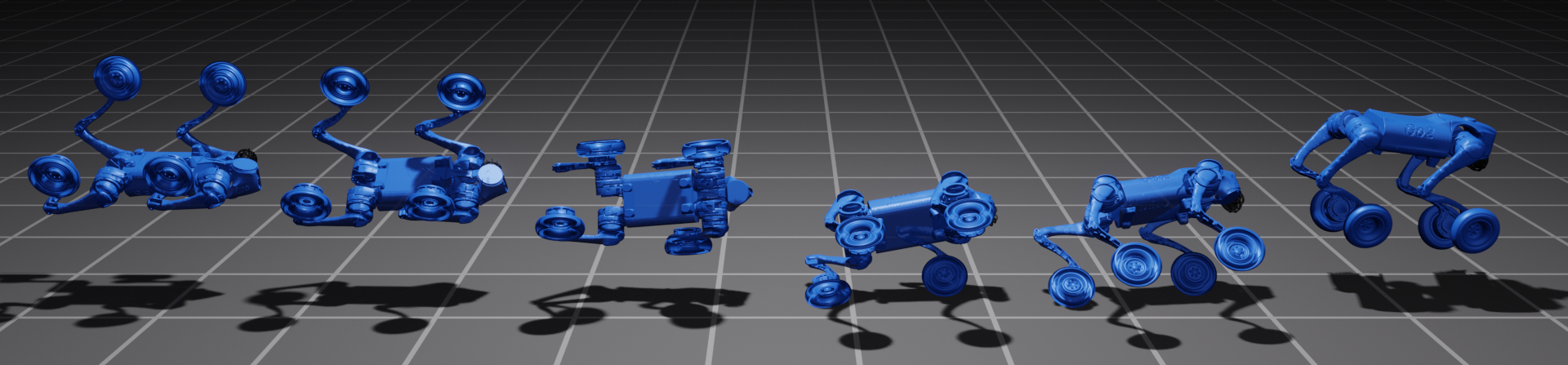}
        \caption{\scriptsize Backward lateral locomotion with cartwheel-style leg rotations and circular trajectories.}
        \label{fig:skill_backward_lateral}
    \end{subfigure}


    \begin{subfigure}[t]{0.98\linewidth}

        \centering
        \includegraphics[width=0.98\linewidth, trim={0 1cm 0 1cm},  
        clip]{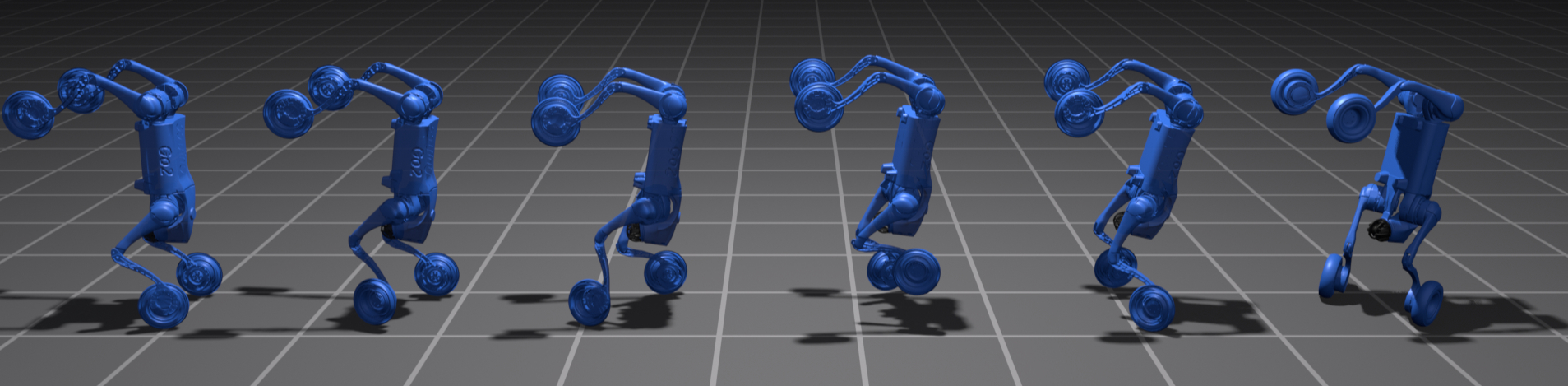}
        \caption{\scriptsize Front legs push-glide-push sequence for skating forward.}
        \label{fig:skill_active_rolling}
    \end{subfigure}

    \caption{\textbf{Qualitative Skill Synthesis Results.} We present representative skills synthesized by MimicAgent. Each subfigure shows a different skill (with the initial prompt in the subcaption); please see the project page for trajectory rollouts.} 
    \label{fig:qualitative}
\end{figure}

\begin{figure}[t]
    \centering
    \begin{subfigure}[t]{0.98\linewidth}
        \centering
        \includegraphics[width=0.98\linewidth, trim={0 0 0 7cm}, clip]{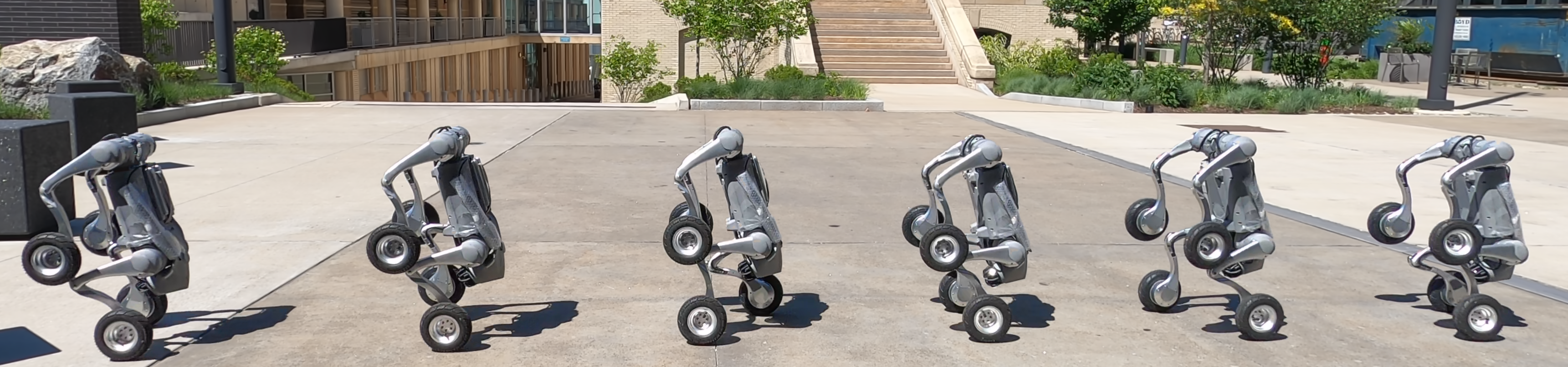}
        \caption{\scriptsize Handstand Walk: Walking at 0.4~m/s on challenging outdoor terrain.}
        \label{fig:hardware_handstand_walk}
    \end{subfigure}

    \vspace{0.15cm}

    \begin{subfigure}[t]{0.98\linewidth}
        \centering
        \includegraphics[width=0.98\linewidth]{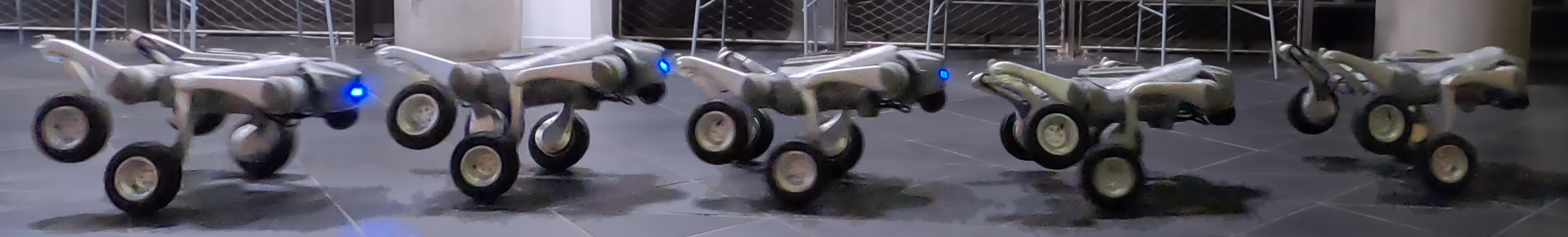}
        \caption{\scriptsize Wheelie Roll: Rolling forward at 0.5 m/s on the front wheels while the rear legs are raised and the torso is parallel to ground}
        \label{fig:hardware_front_wheel_roll}
    \end{subfigure}

    \caption{\textbf{Sim-to-Real Deployment.} MimicAgent policies transfer directly from simulation to the Go2-W quadruped for both (a) two-legged handstand walking and (b) rolling on the front wheels; please see the project page for rollouts.}
    \vspace{-5.25mm}
    \label{fig:hardware_result}
\end{figure}

\textbf{Qualitative Results.} We visualize  policies trained with MimicAgent reference trajectories in simulation (Figure \ref{fig:qualitative}) and real world deployments (Figure \ref{fig:hardware_result}). We find that MimicAgent can successfully learn dynamic skills in simulation, and highlight that MimicAgent can generate (to the best of our knowledge) never-before-seen skating behavior. Although the reference trajectory does not capture the nuances of skating on underactuated wheels while balancing on two legs, training a policy with DeepMimic \textit{does} capture such subtle dynamics. In Figure \ref{fig:hardware_result}, we show that walking on two legs is actually transferable to a real robot. We stress that the reference trajectory generated by MimicAgent is not stable, yet the learned RL policy is able to maintain balance while walking.

\textbf{Sim-to-Real Metrics.} We evaluate 5 skills on the Go2-W over 5 trials each on flat ground of varying roughness (concrete, mat, and polished floor), following the metric conventions of RMA~\cite{kumar2021rma} (Table ~\ref{tab:skill_robustness}). A trial succeeds only if the robot completes $90\%$ of an episode without an unintended body part touching the ground and matches the high level skill description. We find that Trot and Handstand Walk succeed in all 5 trials at every roughness level, reaching $1.00$ normalized time-to-failure (TTF) and distance traveled (DT). Handstand Skate fails a single trial, losing balance while dragging its feet during start-up on rough ground, leaving both TTF and DT at $0.84$. Wheelie Roll likewise fails one trial from premature rear-foot contact ($80\%$ success). Notably, its DT stays at $0.95$ while TTF drops to $0.91$; this gap can be attributed to Roll failing on contact violations within an otherwise completed rollout. Lastly, Spin Transition, in which the robot spins in place on all 4 legs and slowly transitions to spinning on its 2 front legs, passes every trial with clean lift-off during the transition phase. 

\begin{table}[t]
\vspace{2mm}
\centering
\footnotesize
\setlength{\tabcolsep}{4pt}
\setlength{\abovecaptionskip}{4pt}
\caption{\textbf{Sim-to-Real Robustness.} We evaluate the success rate and policy rollout quality across five skills.
}
\label{tab:skill_robustness}
\renewcommand{\arraystretch}{1.15}
\begin{tabular*}{\columnwidth}{@{\extracolsep{\fill}}l|ccc@{}}
\toprule
\textbf{Skill}
  & \cellcolor{blue!8}\makecell{\textbf{SR} $\uparrow$}
  & \cellcolor{blue!8}\makecell{\textbf{TTF} $\uparrow$}
  & \cellcolor{blue!8}\makecell{\textbf{DT} $\uparrow$} \\
\midrule
 Trot            & \textbf{100\% (5/5)} & \textbf{1.00} & \textbf{1.00} \\
 Handstand Walk  & \textbf{100\% (5/5)} & \textbf{1.00} & \textbf{1.00} \\
 Handstand Skate & 80\% (4/5)           & 0.84          & 0.84 \\
 Wheelie Roll         & 80\% (4/5)           & 0.91          & 0.95 \\
 Spin Transition            & \textbf{100\% (5/5)}         & \textbf{1.00}        & N/A \\
\bottomrule
\end{tabular*}
\vspace{-6mm}
\end{table}

\textbf{Analysis of Failure Cases.} 
We find that MimicAgent reaches a downstream policy success rate of $92\%$ across $50$ generated skills (Table ~\ref{tab:human_scores_avg}). To isolate the effect of reference-trajectory quality on downstream policy performance, we perform a failure-case analysis on the four remaining skills. All four are aerial or in-place rotational skills. Here, two properties coincide. First, tracking objectives of the DeepMimic family are defined primarily on root-relative end-effector positions, so for this skill class a stationary counterfactual that reproduces the reference joint trajectory retains $96.5\text{--}99.3\%$ of the perfect-tracking reward. Second, MimicAgent's trajectory quality degrades on a subset of these failures, producing flight phases that do not follow free-fall motion. Specifically, the robot rises slowly and stays airborne far longer than is physically realizable. We posit that these failures can be attributed to both, since a reference the robot cannot physically track leaves a counterfactual motion that is nearly as rewarding as the intended skill.

\section{Conclusion}
In this paper, we present MimicAgent, an LLM-based agentic pipeline for quadruped trajectory generation. By leveraging coding agents to synthesize coarse yet kinematically feasible reference motions, MimicAgent sidesteps the brittleness and manual effort of reward shaping while avoiding the need for large-scale motion capture datasets. Our results show that these automatically generated trajectories are sufficient for training robust and expressive behaviors across a diverse set of skills. Human evaluations consistently favor policies trained with MimicAgent over 
Eureka, highlighting our method's motion quality and language following abilities.

\bibliographystyle{IEEEtran}
\bibliography{egbib.bib}

@article{ma2023eureka,
  title={Eureka: Human-level reward design via coding large language models},
  author={Ma, Yecheng Jason and Liang, William and Wang, Guanzhi and Huang, De-An and Bastani, Osbert and Jayaraman, Dinesh and Zhu, Yuke and Fan, Linxi and Anandkumar, Anima},
  journal={arXiv:2310.12931},
  year={2023}
}

@inproceedings{margolis2023walk,
  title={Walk these ways: Tuning robot control for generalization with multiplicity of behavior},
  author={Margolis, Gabriel B and Agrawal, Pulkit},
  booktitle={Conference on Robot Learning},
  pages={22--31},
  year={2023},
  organization={PMLR}
}

@article{kumar2021rma,
  title={Rma: Rapid motor adaptation for legged robots},
  author={Kumar, Ashish and Fu, Zipeng and Pathak, Deepak and Malik, Jitendra},
  journal={arXiv:2107.04034},
  year={2021}
}

@article{lee2020learning,
  title={Learning quadrupedal locomotion over challenging terrain},
  author={Lee, Joonho and Hwangbo, Jemin and Wellhausen, Lorenz and Koltun, Vladlen and Hutter, Marco},
  journal={Science robotics},
}

@article{nahrendra2023dreamwaq,
  title={Dreamwaq: Learning robust quadrupedal locomotion with implicit terrain imagination via deep reinforcement learning},
  author={Nahrendra, I and Yu, Byeongho and Myung, Hyun},
  journal={arXiv:2301.10602},
  year={2023}
}

@article{allshire2025visual,
  title={Visual Imitation Enables Contextual Humanoid Control},
  author={Allshire, Arthur and Choi, Hongsuk and Zhang, Junyi and McAllister, David and Zhang, Anthony and Kim, Chung Min and Darrell, Trevor and Abbeel, Pieter and Malik, Jitendra and Kanazawa, Angjoo},
  journal={arXiv:2505.03729},
  year={2025}
}

@article{peng2018deepmimic,
  title={Deepmimic: Example-guided deep reinforcement learning of physics-based character skills},
  author={Peng, Xue Bin and Abbeel, Pieter and Levine, Sergey and Van de Panne, Michiel},
  journal={ACM Transactions On Graphics (TOG)},
  year={2018},
}

@inproceedings{xu2025intermimic,
  title={Intermimic: Towards universal whole-body control for physics-based human-object interactions},
  author={Xu, Sirui and Ling, Hung Yu and Wang, Yu-Xiong and Gui, Liang-Yan},
  booktitle={Proceedings of the Computer Vision and Pattern Recognition Conference},
  pages={12266--12277},
  year={2025}
}

@article{yu2023language,
  title={Language to rewards for robotic skill synthesis},
  author={Yu, Wenhao and Gileadi, Nimrod and Fu, Chuyuan and Kirmani, Sean and Lee, Kuang-Huei and Arenas, Montse Gonzalez and Chiang, Hao-Tien Lewis and Erez, Tom and Hasenclever, Leonard and Humplik, Jan and others},
  journal={arXiv:2306.08647},
  year={2023}
}

@conference{AMASS:ICCV:2019,
  title = {{AMASS}: Archive of Motion Capture as Surface Shapes},
  author = {Mahmood, Naureen and Ghorbani, Nima and Troje, Nikolaus F. and Pons-Moll, Gerard and Black, Michael J.},
  booktitle = {International Conference on Computer Vision},
  pages = {5442--5451},
  month = oct,
  year = {2019},
  month_numeric = {10}
}

@inproceedings{cui2024anyskill,
  title={Anyskill: Learning Open-Vocabulary Physical Skill for Interactive Agents},
  author={Cui, Jieming and Liu, Tengyu and Liu, Nian and Yang, Yaodong and Zhu, Yixin and Huang, Siyuan},
  booktitle={Conference on Computer Vision and Pattern Recognition(CVPR)},
  year={2024}
}

@inproceedings{cui2025grove,
  title={Grove: A generalized reward for learning open-vocabulary physical skill},
  author={Cui, Jieming and Liu, Tengyu and Meng, Ziyu and Yu, Jiale and Song, Ran and Zhang, Wei and Zhu, Yixin and Huang, Siyuan},
  booktitle={Proceedings of the Computer Vision and Pattern Recognition Conference},
  pages={15781--15790},
  year={2025}
}

@inproceedings{li2024learning,
  title={Learning agile bipedal motions on a quadrupedal robot},
  author={Li, Yunfei and Li, Jinhan and Fu, Wei and Wu, Yi},
  booktitle={2024 IEEE International Conference on Robotics and Automation (ICRA)},
  pages={9735--9742},
  year={2024},
  organization={IEEE}
}

@article{zhou2025adaptive,
  title={Adaptive Interactive Navigation of Quadruped Robots using Large Language Models},
  author={Zhou, Kangjie and Mu, Yao and Song, Haoyang and Zeng, Yi and Wu, Pengying and Gao, Han and Liu, Chang},
  journal={arXiv:2503.22942},
  year={2025}
}

@article{jian2025lapp,
  title={LAPP: Large Language Model Feedback for Preference-Driven Reinforcement Learning},
  author={Jian, Pingcheng and Wei, Xiao and Liu, Yanbaihui and Moore, Samuel A and Zavlanos, Michael M and Chen, Boyuan},
  journal={arXiv:2504.15472},
  year={2025}
}

@inproceedings{qiu2025wildlma,
  title={Wildlma: Long horizon loco-manipulation in the wild},
  author={Qiu, Ri-Zhao and Song, Yuchen and Peng, Xuanbin and Suryadevara, Sai Aneesh and Yang, Ge and Liu, Minghuan and Ji, Mazeyu and Jia, Chengzhe and Yang, Ruihan and Zou, Xueyan and others},
  booktitle={ICRA},
  year={2025},
  organization={IEEE}
}

@article{smith2023learning,
  title={Learning and adapting agile locomotion skills by transferring experience},
  author={Smith, Laura and Kew, J Chase and Li, Tianyu and Luu, Linda and Peng, Xue Bin and Ha, Sehoon and Tan, Jie and Levine, Sergey},
  journal={arXiv:2304.09834},
  year={2023}
}

@article{peng2018sfv,
  title={Sfv: Reinforcement learning of physical skills from videos},
  author={Peng, Xue Bin and Kanazawa, Angjoo and Malik, Jitendra and Abbeel, Pieter and Levine, Sergey},
  journal={ACM Transactions On Graphics (TOG)},
  volume={37},
  number={6},
  pages={1--14},
  year={2018},
  publisher={ACM New York, NY, USA}
}

@inproceedings{he2024learning,
  title={Learning human-to-humanoid real-time whole-body teleoperation},
  author={He, Tairan and Luo, Zhengyi and Xiao, Wenli and Zhang, Chong and Kitani, Kris and Liu, Changliu and Shi, Guanya},
  booktitle={2024 IEEE/RSJ International Conference on Intelligent Robots and Systems (IROS)},
  pages={8944--8951},
  year={2024},
  organization={IEEE}
}

@INPROCEEDINGS{cheetah3,
  author={Di Carlo, Jared and Wensing, Patrick M. and Katz, Benjamin and Bledt, Gerardo and Kim, Sangbae},
  booktitle={2018 IEEE/RSJ International Conference on Intelligent Robots and Systems (IROS)}, 
  title={Dynamic Locomotion in the MIT Cheetah 3 Through Convex Model-Predictive Control}, 
  year={2018},
  volume={},
  number={},
  pages={1-9},
  doi={10.1109/IROS.2018.8594448}}

@inproceedings{grandia2019feedback,
  title={Feedback mpc for torque-controlled legged robots},
  author={Grandia, Ruben and Farshidian, Farbod and Ranftl, Ren{\'e} and Hutter, Marco},
  booktitle={2019 IEEE/RSJ International Conference on Intelligent Robots and Systems (IROS)},
  year={2019},
}

@article{RAIBERT199079,
title = {Trotting, pacing and bounding by a quadruped robot},
journal = {Journal of Biomechanics},
year = {1990},
author = {Marc H. Raibert},
}

@INPROCEEDINGS{eckert_bounding,
  author={Eckert, Peter and Spröwitz, Alexander and Witte, Hartmut and Ijspeert, Auke Jan},
  booktitle={International Conference on Robotics and Automation}, 
  title={Comparing the effect of different spine and leg designs for a small bounding quadruped robot}, 
  year={2015},
  }

@ARTICLE{1255398,
  author={Marhefka, D.W. and Orin, D.E. and Schmiedeler, J.P. and Waldron, K.J.},
  journal={Transactions on Mechatronics}, 
  title={Intelligent control of quadruped gallops}, 
  year={2003},
    }

@article{Kress-Gazit01012008,
author = {Hadas Kress-Gazit and Georgios E. Fainekos and George J. Pappas},
title = {Translating Structured English to Robot Controllers},
journal = {Advanced Robotics},
year = {2008},
}

@article{brohan2022rt,
  title={Rt-1: Robotics transformer for real-world control at scale},
  author={Brohan, Anthony and Brown, Noah and Carbajal, Justice and Chebotar, Yevgen and Dabis, Joseph and Finn, Chelsea and Gopalakrishnan, Keerthana and Hausman, Karol and Herzog, Alex and others},
  journal={arXiv:2212.06817},
  year={2022}
}

@article{mees2022matters,
  title={What matters in language conditioned robotic imitation learning over unstructured data},
  author={Mees, Oier and Hermann, Lukas and Burgard, Wolfram},
  journal={IEEE Robotics and Automation Letters},
  year={2022},
  publisher={IEEE}
}

@article{liang2022code,
  title={Code as policies: Language model programs for embodied control},
  author={Liang, Jacky and Huang, Wenlong and Xia, Fei and Xu, Peng and Hausman, Karol and Ichter, Brian and Florence, Pete and Zeng, Andy},
  journal={arXiv:2209.07753},
  year={2022}
}

@article{singh2022progprompt,
  title={Progprompt: Generating situated robot task plans using large language models},
  author={Singh, Ishika and Blukis, Valts and Mousavian, Arsalan and Goyal, Ankit and Xu, Danfei and Tremblay, Jonathan and Fox, Dieter and Thomason, Jesse and Garg, Animesh},
  journal={arXiv:2209.11302},
  year={2022}
}

@article{lin2023text2motion,
  title={Text2motion: From natural language instructions to feasible plans},
  author={Lin, Kevin and Agia, Christopher and Migimatsu, Toki and Pavone, Marco and Bohg, Jeannette},
  journal={Autonomous Robots},
  volume={47},
  number={8},
  pages={1345--1365},
  year={2023},
  publisher={Springer}
}

@inproceedings{bellegarda2025allgaits,
  title={Allgaits: Learning all quadruped gaits and transitions},
  author={Bellegarda, Guillaume and Shafiee, Milad and Ijspeert, Auke},
  booktitle={ICRA},
  year={2025},
  organization={IEEE}
}

@article{zhang2025add,
  title={ADD: Physics-Based Motion Imitation with Adversarial Differential Discriminators},
  author={Zhang, Ziyu and Bashkirov, Sergey and Yang, Dun and Taylor, Michael and Peng, Xue Bin},
  journal={arXiv:2505.04961},
  year={2025}
}

@article{turcato2025towards,
  title={Towards autonomous reinforcement learning for real-world robotic manipulation with large language models},
  author={Turcato, Niccol{\`o} and Iovino, Matteo and Synodinos, Aris and Dalla Libera, Alberto and Carli, Ruggero and Falco, Pietro},
  journal={IEEE Robotics and Automation Letters},
  year={2025},
  publisher={IEEE}
}

@InProceedings{pmlr-v15-ross11a,
  title = 	 {A Reduction of Imitation Learning and Structured Prediction to No-Regret Online Learning},
  author = 	 {Ross, Stephane and Gordon, Geoffrey and Bagnell, Drew},
  booktitle = 	 {AISTATS},
}

\end{document}